\documentclass[journal]{IEEEtran}
\usepackage{amsmath,amsfonts}
\usepackage{algorithmic}
\usepackage{array}
\usepackage[caption=false,font=footnotesize,labelfont=rm,textfont=rm]{subfig}
\usepackage{textcomp}
\usepackage{stfloats}
\usepackage{url}
\usepackage{verbatim}
\usepackage{graphicx}
\usepackage{multirow}
\usepackage{booktabs}
\usepackage{enumerate}
\usepackage{amssymb}
\usepackage[ruled,longend,linesnumbered]{algorithm2e}
\def\BibTeX{{\rm B\kern-.05em{\sc i\kern-.025em b}\kern-.08em
    T\kern-.1667em\lower.7ex\hbox{E}\kern-.125emX}}
\usepackage[colorlinks,linkcolor=red,anchorcolor=blue,citecolor=blue,urlcolor=blue]{hyperref}
\usepackage{balance}

\begin{document}
\title{CLIP-SENet: CLIP-based Semantic Enhancement Network for Vehicle Re-identification}
\author{Liping Lu, 
Zihao Fu, 
Duanfeng Chu,~\IEEEmembership{Member,~IEEE,} 
Wei Wang, and
Bingrong Xu,~\IEEEmembership{Member,~IEEE}
\thanks{This work is supported in part by the National Key Research and Development Program of China (2021YFB2501104), the National Natural Science
Foundation of China under Grant (62206204), and the Natural Science Foundation of Hubei Province, China (2023AFB705). (\textit{Corresponding author: Bingrong Xu.})}
\thanks{
Liping Lu and Zihao Fu are with the School of Computer Science and Artificial Intelligence, Wuhan University of Technology, Wuhan 430063, China (e-mail:
luliping@whut.edu.cn; fu032@whut.edu.cn).

Duanfeng Chu is with the Intelligent
Transportation Systems Research Center, Wuhan University of Technology,
Wuhan 430063, China (e-mail: chudf@whut.edu.cn).

Wei Wang is with the School of Cyber Science and Technology, Shenzhen Campus of Sun Yat-sen University, Shenzhen, 518107, China. (e-mail: wangwei29@mail.sysu.edu.cn).

Bingrong Xu is with the School of Automation, Wuhan
University of Technology, Wuhan 430070, China (e-mail: bingrongxu@whut.edu.cn).
}
}

\markboth{IEEE Transactions on Intelligent Transportation Systems
}%
{Liping Lu, Zihao Fu, Duanfeng Chu, Wei Wang, and Bingrong Xu, \MakeLowercase{\textit{(Lu et al.)}}:
CLIP-SENet: CLIP-based Semantic Enhancement Network for Vehicle Re-identification}

\maketitle

\begin{abstract}
Vehicle re-identification (Re-ID) is a crucial task in intelligent transportation systems (ITS), aimed at retrieving and matching the same vehicle across different surveillance cameras. Numerous studies have explored methods to enhance vehicle Re-ID by focusing on semantic enhancement. However, these methods often rely on additional annotated information to enable models to extract effective semantic features, which brings many limitations.
In this work, we propose a CLIP-based Semantic Enhancement Network (CLIP-SENet), an end-to-end framework designed to autonomously extract and refine vehicle semantic attributes, facilitating the generation of more robust semantic feature representations. Inspired by zero-shot solutions for downstream tasks presented by large-scale vision-language models, we leverage the powerful cross-modal descriptive capabilities of the CLIP image encoder to initially extract general semantic information. Instead of using a text encoder for semantic alignment, we design an adaptive fine-grained enhancement module (AFEM) to adaptively enhance this general semantic information at a fine-grained level to obtain robust semantic feature representations. These features are then fused with common Re-ID appearance features to further refine the distinctions between vehicles.
Our comprehensive evaluation on three benchmark datasets demonstrates the effectiveness of CLIP-SENet. Our approach achieves new state-of-the-art performance, with 92.9\% mAP and 98.7\% Rank-1 on VeRi-776 dataset, 90.4\% Rank-1 and 98.7\% Rank-5 on VehicleID dataset, and 89.1\% mAP and 97.9\% Rank-1 on the more challenging VeRi-Wild dataset.
\end{abstract}


\begin{figure*}[!t]
\centering
\includegraphics[width=\linewidth]{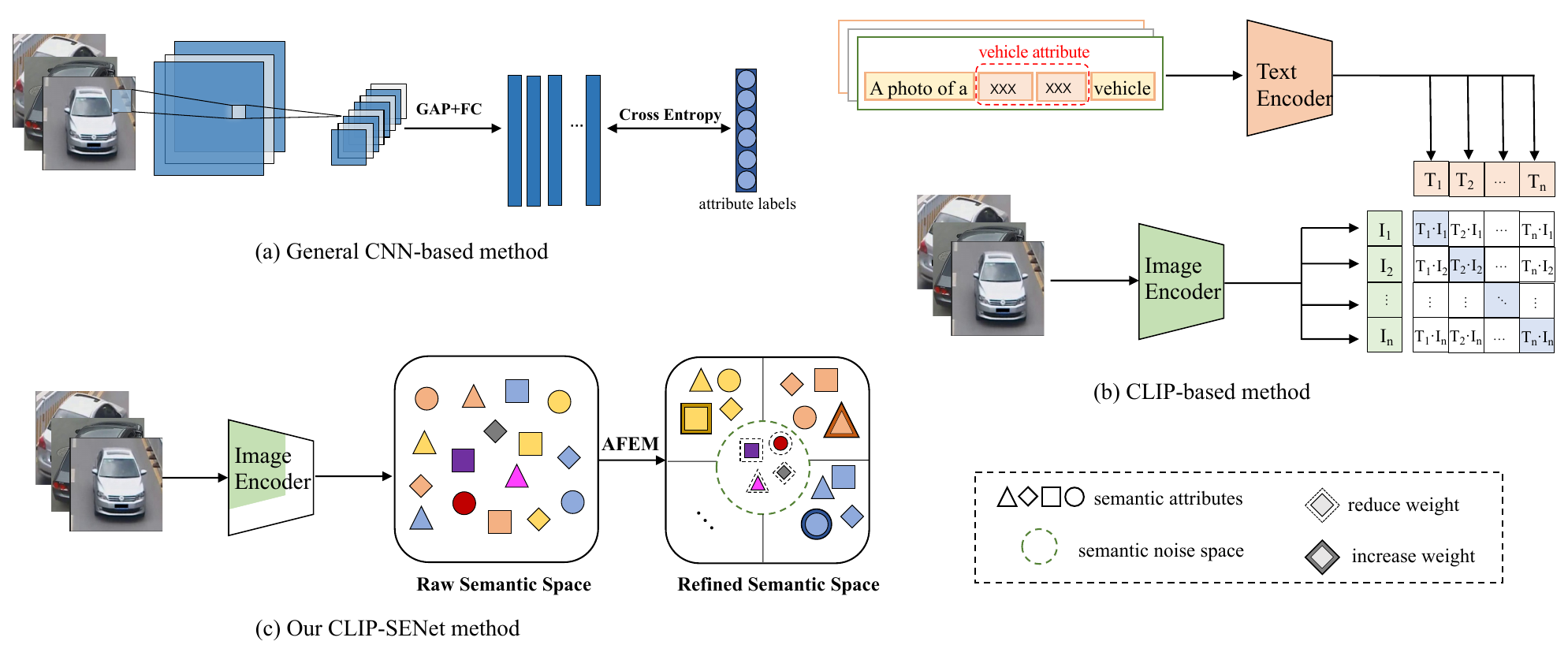}
\caption{Comparison of semantic enhancement methods in Re-ID.
The semantic noise represents features that are not strongly correlated with the semantic attributes of vehicles, such as background roads and irrelevant obstructions.
Increasing and decreasing weights are used to abstractly demonstrate the adjustments of different semantic features weights during model training.
} 
\label{fig:intro}
\end{figure*}

\section{Introduction}
\IEEEPARstart{V}{ehicle} re-identification (Re-ID) is a pivotal task in transportation surveillance and security systems, aiming to accurately retrieve and match vehicle images captured by disparate surveillance cameras. 
The rapid development of deep convolutional neural networks (CNNs) has led to significant breakthroughs in vehicle Re-ID efforts~\cite{luo2019bag,he2023fastreid,He_2021_ICCV,huynh2021strong}.
Despite these advancements, vehicle Re-ID still faces challenges. The significant appearance variations in images of the same vehicle, caused by changes in camera angles and lighting conditions, result in substantial intra-class differences.
Therefore, modeling viewpoint-invariant features of vehicle appearance has become a key strategy for vehicle Re-ID.

When considering invariant features, vehicle attributes such as type, color, and brand easily come to mind. These attributes can serve as robust markers for vehicle identification, providing a stable foundation for Re-ID under varying conditions. Previous attribute enhancement methods in Re-ID~\cite{zhao2019structural,qian2020stripe, 9106791,li2022attribute,yu2023semantic} focus on highlighting different vehicle attributes by utilizing annotated information from datasets when extracting appearance features. However, this approach, which relies on attribute features learned through the cross-entropy loss function, has two main drawbacks.
First, not all datasets provide annotated attribute information for each vehicle, making this attribute extraction method inherently inefficient and lacking in generalization capability. It is constrained by the availability and quality of the annotated information. Second, these methods often fail to effectively balance the weight of attribute features and appearance features in the final feature representation. They highlight some attributes information of appearance features, leading to a bias towards attribute classification and overlooking important fine-grained details in the appearance. This poses a significant challenge for classifying vehicle IDs with small inter-class differences.

The emergence of large-scale visual-language models has provided a novel perspective for the extraction of semantic features. Among these, the Contrastive Language-Image Pre-training (CLIP \cite{radford2021learning}) framework stands out for its innovative approach to cross-modal representation learning. By training on a broad array of image-text pairs, CLIP effectively embeds images and texts into a cohesive semantic space, enabling the alignment of visual features with textual descriptions and thus enriching the extracted features with substantial semantic content. These features are transferable and adaptable to a multitude of different tasks. The influence of CLIP has propelled a series of investigative efforts~\cite{li2023clip, lin2023exploring, yan2023clip} within the Re-ID research community, aiming to enhance the unique representation of targets by aligning semantic image features with corresponding textual descriptions.
Notably, CLIP-ReID~\cite{li2023clip} employs a two-stage training approach to apply the CLIP paradigm to Re-ID tasks, promoting a significant stride in the field. 
In the initial phase, this method learns unique linguistic descriptions for each vehicle ID. Subsequently, in the second phase, it fine-tunes the image encoder to semantically align image features with these learned language descriptions. Although CLIP-ReID has achieved impressive experimental results, it still has certain limitations. Firstly, while CLIP-ReID does not use annotated attribute information from the dataset but instead generates sentence descriptions using NLP methods, the final representations heavily rely on the quality of the generated sentences and significantly increase the training time of model. Secondly, the two-stage learning process introduces complexity in parameter tuning and model convergence during training and fine-tuning of the large model.
\IEEEpubidadjcol

In this paper, we propose the CLIP-based Semantic Enhancement Network (CLIP-SENet), which can efficiently extract vehicle semantic attributes and enhance the unique representation of vehicle features. Unlike previous CNN-based and CLIP-Based methods, detailed in Fig.~\ref{fig:intro}, our approach does not require any additional textual annotation information for training to extracted semantic features. 
CLIP-SENet not only discarded the text encoder from CLIP but also used a lightweight CLIP model obtained through knowledge distillation, known as TinyCLIP~\cite{wu2023tinyclip}, to eliminate the dependency on textual annotations and reduce the complexity of tuning large models. We utilized the image encoder from TinyCLIP to efficiently capture the raw semantic features from images. However, due to the absence of text-semantic alignment, the extracted features contained some noise that affected the Re-ID process. To address this issue, we proposed the Adaptive Fine-Grained Enhancement Module (AFEM), which employs adaptive weighting to filter the raw semantic information, reducing noise and emphasizing attributes that are beneficial for distinguishing different vehicles. Eventually, we  combine the refine features with appearance features extracted by CNN baseline to strengthen the final features representation, resulting in more robust and accurate vehicle Re-ID performance.

In summary, the contributions of this paper can be outlined in the following aspects:
\begin{itemize}
\item {We propose the CLIP-based Semantic Enhancement Network (CLIP-SENet) to efficiently extract vehicle semantic information in an unsupervised manner, offering a novel perspective for the attribute-based enhancement method in vehicle Re-ID.} 
\item {We investigated the performance of using the CLIP image encoder alone for semantic extraction in downstream tasks, further demonstrating the powerful cross-modal semantic representation capabilities of the CLIP model.}
\item {We propose the Adaptive Fine-grained Enhancement Module (AFEM) to refine the raw semantic information extracted by the image encoder from TinyCLIP.  
The AFEM applies adaptive weighting to underscore essential semantic details while minimizing the impact of irrelevant semantic information.}
\item {Extensive experiments on three benchmark datasets in vehicle Re-ID, CLIP-SENet demonstrates state-of-the-art performance, surpassing the previous leading method.}
\end{itemize}

\section{Related Work}
\subsection{Attribute-based Enhancement Methods.}
The purpose of Re-ID task is to match objects from a gallery, but there are differences for the matching of different objects (e.g., vehicles and people respectively). Due to the rich attribute information in pedestrian images, some attribute-based feature enhancement was initially popularized in person Re-ID ~\cite{tay2019aanet,yin2023efficient, 10555193,lin2019improving, 9733175}.
Recent works~\cite{quispe2021attributenet,li2022attribute,zhang2022graph,yu2022multi,yu2023semantic} have demonstrated the effectiveness of attribute-based enhancement in vehicle Re-ID. Quispe et al.~\cite{quispe2021attributenet} proposed AttributeNet (ANet) to refine valuable attribute features in Re-ID and integrating them with general ReID features to increase the discrimination capabilities. Li et al.~\cite{li2022attribute} designed the Attribute and State Guided Structural Embedding Network (ASSEN) to enhance the discriminative feature by mitigating the negative effects of illumination and view angle while capitalizing on the positive attributes of color and type.  
Yu et al.~\cite{yu2022multi} use transformer~\cite{vaswani2017attention} for attribute extraction and leverage a multi-attribute adaptive aggregation network to highlight the significance of key attributes. 
However, these attribute-based enhancement works, without exception, require additional annotation information to complete the training in a supervised manner.
In fact, the majority of Re-ID datasets lack attribute labels, and manually annotating attribute information is an exceptionally costly and time-consuming task.

\subsection{Fine-grained Enhancement Methods.}
The Re-ID task can be regarded as a form of fine-grained identification, primarily focusing on discerning intra-class variances while also distinguishing between different target classes. The attention mechanism, such as self-attention, plays a pivotal role in fine-grained recognition. Attention-based fine-grained enhancement has proven effective in numerous Re-ID studies~\cite{rao2021counterfactual,hong2021fine,yin2020fine,zhu2022dual}. Rao et al.~\cite{rao2021counterfactual} proposed a counterfactual attention learning method that leverages causal inference to enhance the effective learning of fine-grained features. 
Hong et al.~\cite{hong2021fine} designed the Shape-Appearance Mutual learning framework (FSAM), in which shape flow and appearance flow complement each other to extract fine body shape features guided by identity. 
Yin et al.~\cite{yin2020fine} used the attention module to focus on pedestrian pose features, which are more unique and help distinguish similar appearances between people.
In contrast to prior efforts, our approach leverages fully connected layer grouping to learn the weights of different semantic attributes, achieving an effect similar to the attention mechanism, but avoid the computational complexity associated with attention module.

\subsection{Large-scale Visual-language Learning.}
In recent years, large-scale visual-language models~\cite{jia2021scaling, li2022blip, li2021align, radford2021learning} have gained widespread popularity, with CLIP~\cite{radford2021learning} being a prominent example and serving as the foundation for numerous derivative works. CLIP is a pre-trained model based on contrastive text-image pairs, capable of understanding the content in images and associating it with textual descriptions. It exhibits strong generality and achieves impressive zero-shot performance in downstream tasks. Nonetheless, the reliance of CLIP on larger model capacities presents computational efficiency challenges, serving as a prominent impediment to its practical deployment. Notably, training smaller models directly often leads to suboptimal performance, necessitating compression techniques to yield more compact and faster models without compromising their effectiveness.
To address this problem, Wu et al.~\cite{wu2023tinyclip} introduced TinyCLIP, a novel method using knowledge distillation to compress the CLIP model. TinyCLIP effectively compresses the parameters of the CLIP model through affinity mimicry and weight inheritance, maintaining a lightweight size while demonstrating remarkable zero-shot accuracy on ImageNet with minimal parameters and exhibiting strong transfer capability to downstream tasks.
Considering the comprehensive aspects of our proposed model, we opted for image encoder from TinyCLIP for our experiments to alleviate the burden of model parameters.

\subsection{CLIP-based Re-ID Methods.}
With the popularity of CLIP framework, the Re-ID field has seen numerous investigations~\cite{li2023clip, lin2023exploring, yan2023clip} into applying the CLIP framework to Re-ID tasks. The pioneering CLIP-ReID first applied the CLIP paradigm to Re-ID tasks in a two-stage training approach, achieving highly competitive results. Given the lack of specific textual descriptions for target IDs in Re-ID datasets, traditional text-image contrastive learning proves difficult to implement. CLIP-ReID~\cite{li2023clip}, via the CoOp~\cite{zhou2022learning} method, generates textual prompts for each target in the first stage, fine-tuning the image encoder in the second stage to adapt to Re-ID tasks. The successful application of the CLIP paradigm has significantly advanced the Re-ID field. Yan et al~\cite{yan2023clip} explored the extraction of fine-grained information in pedestrian re-identification with the CLIP model, aiming to leverage robust capabilities of CLIP for cross-modal fine-grained alignment to enhance the performance of Re-ID models. 
From the methods mentioned above, it's clear that applying CLIP paradigm to Re-ID tasks involves training with additional text prompts. However, these trained text prompts can be unstable, significantly increasing the complexity of model training.

\begin{figure*}[!t]
\includegraphics[width=\linewidth]{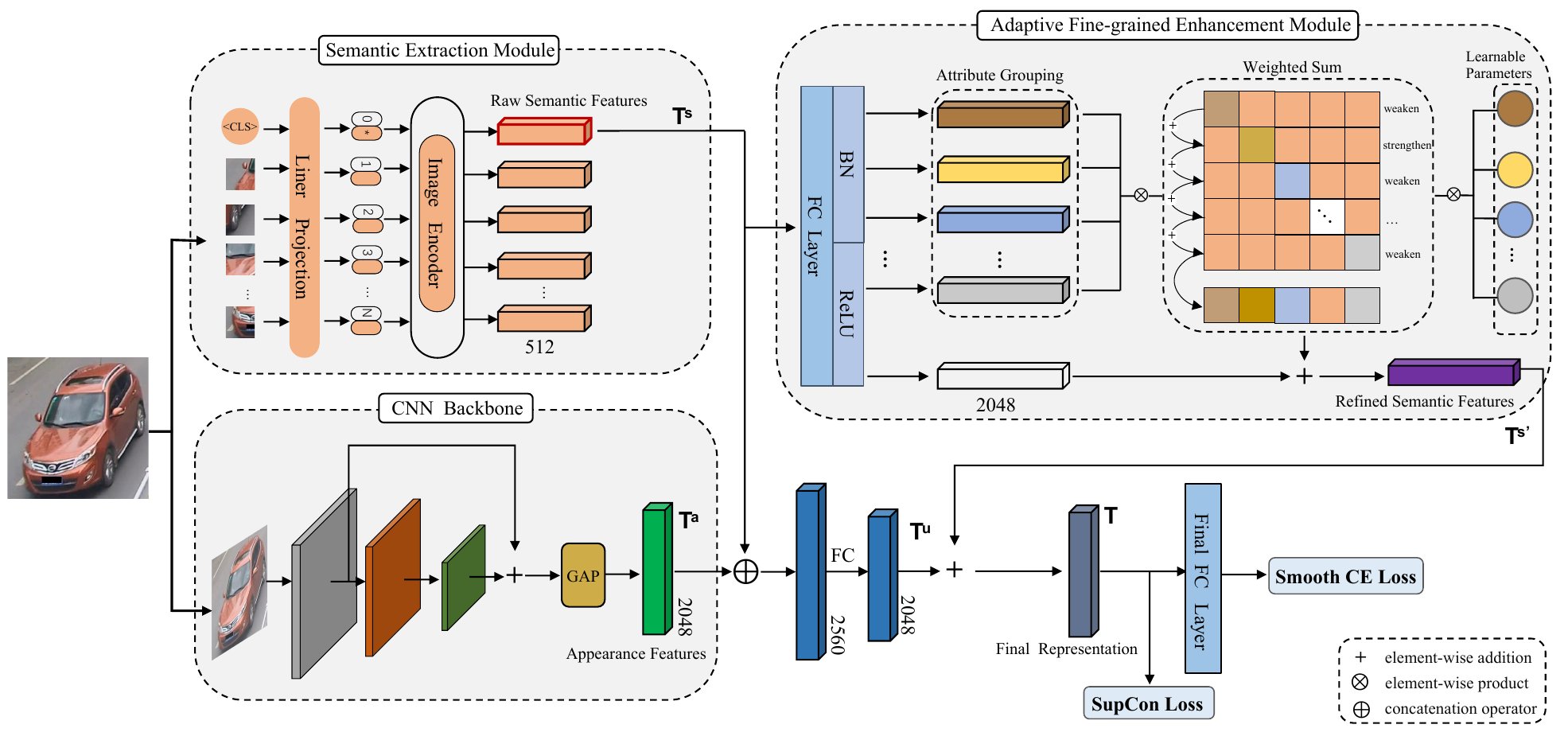}
\caption{The pipeline of the CLIP-SENet framework.
For input images, both the CNN Backbone and the SEM process them in parallel. The CNN Backbone initially processes and extracts appearance features from the images. Concurrently, to prepare the images for SEM, they are processed to fit the input format of the ViT~\cite{dosovitskiy2020image}. Then, the SEM extracts raw semantic embeddings from vehicle images. These semantic embeddings are then fused with the vehicle appearance features in a high-dimensional space to maximize the preservation of semantic information. Meanwhile, the AFEM applies adaptive weighting to these raw semantic features,  reducing the weight of noisy attributes while favoring those conducive to identity identification, resulting in refined semantic attributes. Finally, the refined features are added element-wise with the fused features to enhance the final feature representation.}
\label{fig:model}
\end{figure*}

\section{Methodology}
Vehicle Re-ID is tasked with matching vehicle images from non-overlapping surveillance cameras, identifying those in the gallery matching the identity of a probe vehicle. Traditional methods have struggled to efficiently extract attributes without relying on annotated labels and previous CLIP-based method need addition step for traing a textual description for each vehicle. Addressing this, we introduce the CLIP-based Semantic Enhancement Network (CLIP-SENet), which effectively improves semantic attribute feature extraction from vehicle images, as depicted in Fig.~\ref{fig:model}. CLIP-SENet integrates three key components: the CNN Backbone for initial image processing, the image encoder from TinyCLIP as The Semantic Extraction Module (SEM) for extrating raw semantic attributes, and the Adaptive Fine-grained Enhancement Module (AFEM) for refining these attributes into discriminative features. To strike a balance between vehicle identity accuracy and feature representation similarity during training, we adopt a dual-loss strategy common among Re-ID methodologies. This combines Smooth Cross-Entropy (CE) loss for precise identity classification with Supervised Contrastive (SupCon) loss to enhance the distinction between different vehicle identities.

\subsection{CNN Backbone}
We incorporate the Instance Batch Normalization (IBN) network family~\cite{pan2018two} into our CNN backbone to enhance its performance. The integration of IBN into CNN models like ResNet~\cite{he2016deep}, ResNeXt~\cite{xie2017aggregated}, and SE-Net~\cite{hu2018squeeze} has been shown to effectively improve the modeling of appearance-related features. This is achieved by providing a balance between instance normalization and batch normalization, thereby improving feature generalization across various conditions. We use the ResNeXt with IBN network without final linear classification layer as our CNN backbone, $f_{backbone}(\cdot;\theta)$, to extract vehicle appearance features. For an input batch data  $[X]_N$, the tensor encoded by the network is denoted as:
\begin{equation}
T^{a} = GAP( f_{backbone}([X]_N);\theta) \in \mathbb{R}^{N \times D}
\end{equation}
where  $GAP(\cdot)$ denotes a global average pooling, $N$ and $D$ are batch size and the dimensions of tensor. 

\subsection{Semantic Extraction Module}
Diverging from previous methodologies that rely on dataset annotations to develop semantic extraction capabilities, our SEM directly employs the pre-trained image encoder from TinyCLIP, denoted as $I(\cdot;\varepsilon)$, to extract semantic attributes for the target dataset. This approach notably eliminates the need for any prior annotation information.

In the CLIP paradigm, when fitting the CLIP model to a specific task dataset, the image encoder and text encoder need to work together. Additional text annotations from that dataset are fed into the text encoder to extract semantic vectors, which are then aligned with visual features in a high-dimensional space.  Unlike other visual tasks, vehicle Re-ID datasets often lack the definitive semantic annotations. Previous efforts, such as CLIP-ReID, have addressed the lack of attribute text information in datasets by using NLP techniques to generate learnable text prompts. In contrast, we take an entirely opposite approach by discarding the text encoder altogether and relying solely on a pre-trained image encoder to extract the raw semantic attributes from images. Then, a subsequent attribute fine-grained enhancement network is used to adaptively weight specific semantic features, obtaining more refined semantic features. This setup reduces the parameter count of the CLIP model by half, and we also employ TinyCLIP, a compact model derived from knowledge distillation of CLIP, to further decrease the actual parameter usage of large-scale visual-language models.

In the SEM, the image encoder from TinyCLIP serves as a crucial component for extracting semantic attributes, working in concert with the CNN Backbone to capture an extensive feature set of vehicles. Concurrently, the image encoder processes visual inputs to  transform images into a high-dimensional semantic space, where semantic attributes are encoded as vectors, denoted as:
\begin{equation}
    T^{s} = I([X]_N;\varepsilon)
\end{equation}
This dual process ensures the effective capture of both local appearance features and intrinsic semantic differences of the vehicle.
To effectively amalgamate semantic and appearance features, we concatenate these two distinct types of features and then apply a fully connected (FC) layer. This process yields a 2048-dimensional composite feature representation, $T^{u}$, which retains the nuanced details of vehicle appearance and complete semantic information. The purpose of this fusion is to better update the parameters $\varepsilon$ and $\theta$ during backpropagation. It can be expressed as:
\begin{subequations}
\begin{align}
    T^{u} &= FC[T^{s} \oplus T^{a}]\\
         &= FC[I([X]_N)) \oplus GAP(f_{backbone}([X]_N)] 
\end{align}
\end{subequations}
where $[X]_N$ represent a batch of vehicle images and ``$\oplus$" represents the concatenation operation.

\subsection{Adaptive Fine-grained Enhancement Module}
Despite $T^{s}$ possessing raw semantic information, due to the lack of alignment through a text encoder, it is mixed with extensive noisy semantic information and generic vehicle descriptions, which hinder the ability to distinguish between vehicles with similar appearances but different IDs.
To address this issue, we propose the Adaptive Fine-grained Enhancement Module (AFEM), aimed at separating fine-grained semantic information and suppress noise information for $T^{s}$. 

For fine-grained recognition tasks, a common approach is to use attention mechanisms to focus on key objects or regions in the image, thereby enhancing network performance and discrimination capabilities. 
However, this often leads to a large number of parameters.
To address this issue, we divide the raw semantic features into groups, and by incorporating adaptive parameter learning, we assess the importance of different semantic features for vehicle Re-ID, thereby refining the features. This entire procedure is performed on FC layers, reducing computational overhead.

Specifically, we input $T^{s}$ into a group of linear mapping layer, batch normalization (BN) layer and rectified linear unit (ReLU) layer, producing $G+1$ vectors. These vectors are then split into two branches: one consists of $G$ grouped vectors forming a group-aware representation, which allows a set of learnable weight parameters to independently optimize the information within each group. These learnable parameters are initialized following a standard normal distribution and are updated continuously during backpropagation. The other branch retains the parametric information of the original features. Finally, the aggregated weighted group features are element-wise added to the original features, resulting in the final feature representation, the entire process can be expressed as:
\begin{equation}
    T^{s'} = f_{linear}(T^{s}) + \sum_{i=0}^{G}{w_i \otimes f_{linear}(T^{s})} 
\end{equation}
where $f_{linear}$ includes linear mapping, BN and ReLU processing, $G$ represents the number of groups, ``$\otimes$" denotes element-wise product, $w_i$ represents a set of adaptive learning parameters and ``$+$" denotes element-wise addition.

Through the aforementioned process, AFEM achieves semantic separation by grouping and adaptive adjusting weights of semantic features, distinguishing strongly and weakly correlated attributes for clearer classification. It also reduces weights of non-target features to suppress noise, enhancing model robustness in complex environments. AFEM utilizes a weighting formula to calculate feature group importance and adaptively adjusts these weights based on the loss function.

Finally, we combine the refined semantic features $T^{s'}$ with the fused features $T^{u}$ to obtain the final feature representation, expressed as:
\begin{equation}
    T = T^{u} + T^{s'}
\end{equation}

\begin{algorithm}[!t]
\caption{the training process of CLIP-SENet}
\label{train}
\KwIn{Training dataset $D$, learning rate $\eta$, number of epochs $E$}
\Begin{
\textbf{Initialize} the CNN Backbone, ($f_{backbone}(\cdot;\theta)$), with weights pretrained on ImageNet and the image encoder, $I(\cdot;\varepsilon)$, of TinyCLIP with weights pretrained on LAION and YFCC-400M. Use kaiming initialization for the FC layers.

\textbf{Initialize} Smooth CE Loss as ID loss and initialize SupCon Loss as Metric loss.

\For{$epoch \leftarrow 1$ \KwTo $E$}{
  \For{each batch $[X]_N$ in $D$}{
    $T^{a} \leftarrow  GAP(f_{backbone}([X]_N))$
    
    $T^{s} \leftarrow BN(I([X]_N))$

    $T^{u} \leftarrow FC(T^{u} + T^{s'})$
    
    $T^{s'} \leftarrow f_{linear}(T^{s}) + \sum_{i=0}^{G}{w_i \otimes f_{linear}(T^{s})}$
    
    $T \leftarrow T^{u} + T^{s'}$

    Optimize $\theta$ and $\varepsilon$ by Eq.~\ref{eq_loss}
    
  }
  Update $\eta$ using a cosine annealing scheduler.
}
}
\end{algorithm}

Following most vehicle Re-ID methods, we feed $T$ to a classifier to output the predicted classification $\hat{y}$ and use CE loss function to supervise the ID-related training by comparing $\hat{y}$ with the ground truth label $y$ The formula is as follows:
\begin{small}
\begin{equation}
{L}_{CE} = -\frac{1}{N} \sum_{i=1}^{N} \left[ (1-\varepsilon)\gamma_i\log(\hat{y_i}) + \frac{\varepsilon}{N-1} \sum_{j\neq i} \log(\hat{y_i}) \right]
\end{equation}
\end{small}
where the parameter $\varepsilon\ $, denoting the smoothing factor, is a hyperparameter set to 0.1 in our experiments.

In a training batch, to better enhance the similarity within the same class and increase the dissimilarity between different classes, we use the SupCon loss. The specific formula is as follows:
\begin{equation}
L_{SupCon} = -\frac{1}{N} \sum_{i=1}^{N} \log \frac{e^{\frac{(T_i, T_i^+)}{\tau}}}{e^{\frac{(T_i, T_i^+)}{\tau}} + \sum_{j=1}^{N}e^{\frac{(T_i, T_j^-)}{\tau}}} \quad 
\end{equation}
where $T_i$ represents the features of the anchor sample, ${T}_i^+$ denotes the positive sample (from the same class), and ${T}_j^-$ corresponds to the negative samples. The temperature parameter $\tau$ controls the scale of similarity between feature vectors.

Ultimately, our loss function can be represented as:
\begin{equation}
\label{eq_loss}
L = {L}_{CE} + L_{SupCon}
\end{equation}

By optimizing final representation $T$ through the Eq.~\ref{eq_loss}, we ultimately obtain robust features for vehicle Re-ID.
The entire training process is an end-to-end pipeline, detailed in Algorithm~\ref{train} for comprehensive clarity.

\begin{table*}[t]
\caption{Comparison with state-of-the-art methods on VeRi-776 and VeRi-wild datasets. The superscript star* means that the method incorporates a transformer architecture. Results showcase the top performance in bold for the highest achievement and underscored for the second highest across each metric. }
\label{table:5}
\centering
\renewcommand{\arraystretch}{1.2}
\resizebox{\textwidth}{!}{
\centering
\begin{tabular}{c|c|ccc|cccccc}
\hline
\multirow{2}{*}{Methods} &
\multirow{2}{*}{Reference} &
\multicolumn{3}{c|}{\multirow{2}{*}{VeRi-776}} &
\multicolumn{6}{c}{VeRi-Wild} \\ \cline{6-11} 
 & &
  \multicolumn{3}{c|}{} &
  \multicolumn{2}{c|}{Small} &
  \multicolumn{2}{c|}{Medium} &
  \multicolumn{2}{c}{Large} \\ 
 & &
  \multicolumn{1}{c}{mAP} &
  \multicolumn{1}{c}{Rank-1} &
  Rank-5 &
  \multicolumn{1}{c}{mAP} &
  \multicolumn{1}{c|}{Rank-1} &
  \multicolumn{1}{c}{mAP} &
  \multicolumn{1}{c|}{Rank-1} &
  \multicolumn{1}{c}{mAP} &
  Rank-1 \\ \hline
SAN~\cite{qian2020stripe} & MST 2020 &
  \multicolumn{1}{c}{72.5} &
  \multicolumn{1}{c}{93.3} &
  97.1 &
  \multicolumn{1}{c}{-} &
  \multicolumn{1}{c|}{-} &
  \multicolumn{1}{c}{-} &
  \multicolumn{1}{c|}{-} &
  \multicolumn{1}{c}{-} &
  - \\ 
CAL~\cite{Rao_2021_ICCV} & ICCV 2021 &
  \multicolumn{1}{c}{74.3} &
  \multicolumn{1}{c}{95.4} &
  97.9 &
  \multicolumn{1}{c}{-} &
  \multicolumn{1}{c|}{-} &
  \multicolumn{1}{c}{-} &
  \multicolumn{1}{c|}{-} &
  \multicolumn{1}{c}{-} &
  - \\ 
PVEN~\cite{meng2020parsing} & CVPR 2020 &
  \multicolumn{1}{c}{79.5} &
  \multicolumn{1}{c}{95.6} &
  98.4 &
  \multicolumn{1}{c}{79.8} &
  \multicolumn{1}{c|}{94.0} &
  \multicolumn{1}{c}{73.9} &
  \multicolumn{1}{c|}{92.0} &
  \multicolumn{1}{c}{66.2} &
  88.6 \\ 
DCAL~\cite{zhu2022dual}$^{*}$ & CVPR 2022 &
  \multicolumn{1}{c}{80.2} &
  \multicolumn{1}{c}{96.9} &
  - &
  \multicolumn{1}{c}{-} &
  \multicolumn{1}{c|}{-} &
  \multicolumn{1}{c}{-} &
  \multicolumn{1}{c|}{-} &
  \multicolumn{1}{c}{-} &
  - \\ 
GiT~\cite{shen2023git}$^{*}$ & TIP 2023 &
  \multicolumn{1}{c}{80.3} &
  \multicolumn{1}{c}{96.8} &
  - &
  \multicolumn{1}{c}{81.7} &
  \multicolumn{1}{c|}{92.6} &
  \multicolumn{1}{c}{75.6} &
  \multicolumn{1}{c|}{89.9} &
  \multicolumn{1}{c}{67.5} &
  85.4 \\ 

ANet~\cite{quispe2021attributenet} & IJON 2021 &
  \multicolumn{1}{c}{81.2} &
  \multicolumn{1}{c}{96.8} &
  98.4 &
  \multicolumn{1}{c}{86.9} &
  \multicolumn{1}{c|}{96.5} &
  \multicolumn{1}{c}{82.5} &
  \multicolumn{1}{c|}{95.2} &
  \multicolumn{1}{c}{75.9} &
  92.5 \\ 
ASSEN~\cite{li2022attribute} & TIP 2022 &
  \multicolumn{1}{c}{81.7} &
  \multicolumn{1}{c}{97.3} &
  98.8 &
  \multicolumn{1}{c}{84.3} &
  \multicolumn{1}{c|}{97.1} &
  \multicolumn{1}{c}{78.7} &
  \multicolumn{1}{c|}{\underline{95.6}} &
  \multicolumn{1}{c}{70.1} &
  \underline{93.9} \\ 
FastReID~\cite{he2023fastreid} & MM 2023&
    \multicolumn{1}{c}{81.9} &
    \multicolumn{1}{c}{97.0} &
    \underline{99.0} &
    \multicolumn{1}{c}{{87.7}} &
    \multicolumn{1}{c|}{96.4} &
    \multicolumn{1}{c}{83.5} &
    \multicolumn{1}{c|}{95.1} &
    \multicolumn{1}{c}{77.3} &
    92.5 \\ 
SAVER~\cite{khorramshahi2020devil} & ECCV 2020 &
  \multicolumn{1}{c}{82.0} &
  \multicolumn{1}{c}{96.9} &
  97.7 &
  \multicolumn{1}{c}{80.9} &
  \multicolumn{1}{c|}{94.5} &
  \multicolumn{1}{c}{75.3} &
  \multicolumn{1}{c|}{92.7} &
  \multicolumn{1}{c}{67.7} &
  89.5 \\ 
MsKAT~\cite{9764648}$^{*}$ & TITS 2022 &
  \multicolumn{1}{c}{82.0} &
  \multicolumn{1}{c}{97.1} &
  \underline{99.0} &
  \multicolumn{1}{c}{84.0} &
  \multicolumn{1}{c|}{\underline{97.3}} &
  \multicolumn{1}{c}{78.7} &
  \multicolumn{1}{c|}{\underline{95.6}} &
  \multicolumn{1}{c}{72.2} &
  \underline{93.9} \\   
TransReID~\cite{He_2021_ICCV}$^{*}$ & ICCV 2021 &
  \multicolumn{1}{c}{82.3} &
  \multicolumn{1}{c}{97.1} &
  - &
  \multicolumn{1}{c}{81.2} &
  \multicolumn{1}{c|}{92.3} &
  \multicolumn{1}{c}{-} &
  \multicolumn{1}{c|}{-} &
  \multicolumn{1}{c}{-} &
  - \\ 
HRCN~\cite{zhao2021heterogeneous} & ICCV 2021 &
  \multicolumn{1}{c}{83.1} &
  \multicolumn{1}{c}{97.3} &
  98.9 &
  \multicolumn{1}{c}{85.2} &
  \multicolumn{1}{c|}{94.0} &
  \multicolumn{1}{c}{80.0} &
  \multicolumn{1}{c|}{91.6} &
  \multicolumn{1}{c}{72.2} &
  88.0 \\
vehicleNet~\cite{zheng2020vehiclenet} & TMM 2020 &
  \multicolumn{1}{c}{83.4} &
  \multicolumn{1}{c}{96.7} &
  - &
  \multicolumn{1}{c}{-} &
  \multicolumn{1}{c|}{-} &
  \multicolumn{1}{c}{-} &
  \multicolumn{1}{c|}{-} &
  \multicolumn{1}{c}{-} &
  - \\
CAN\cite{sheng2023discriminative}$^{*}$ & TCSVT 2023 &
  \multicolumn{1}{c}{83.4} &
  \multicolumn{1}{c}{97.6} &
  \underline{99.0} &
  \multicolumn{1}{c}{87.5} &
  \multicolumn{1}{c|}{96.5} &
  \multicolumn{1}{c}{83.2} &
  \multicolumn{1}{c|}{95.2} &
  \multicolumn{1}{c}{77.3} &
  92.5 \\ 
SVRN~\cite{9837795} & TITS 2022 &
  \multicolumn{1}{c}{84.5} &
  \multicolumn{1}{c}{97.2} &
  - &
  \multicolumn{1}{c}{85.5} &
  \multicolumn{1}{c|}{-} &
  \multicolumn{1}{c}{81.5} &
  \multicolumn{1}{c|}{-} &
  \multicolumn{1}{c}{76.3} &
  - \\ 
StrongBaseline~\cite{huynh2021strong} & CVPR 2021 &
  \multicolumn{1}{c}{87.1} &
  \multicolumn{1}{c}{97.1} &
  - &
  \multicolumn{1}{c}{-} &
  \multicolumn{1}{c|}{-} &
  \multicolumn{1}{c}{-} &
  \multicolumn{1}{c|}{-} &
  \multicolumn{1}{c}{-} &
  - \\ 
SOFCT~\cite{yu2023semantic}$^{*}$ & TITS 2024 &
  \multicolumn{1}{c}{80.7} &
  \multicolumn{1}{c}{96.6} &
  98.8 &
  \multicolumn{1}{c}{-} &
  \multicolumn{1}{c|}{-} &
  \multicolumn{1}{c}{-} &
  \multicolumn{1}{c|}{-} &
  \multicolumn{1}{c}{-} &
  - \\
GSE-Net~\cite{10595990} & TITS 2024 &
  \multicolumn{1}{c}{81.3} &
  \multicolumn{1}{c}{96.3} &
  - &
  \multicolumn{1}{c}{-} &
  \multicolumn{1}{c|}{-} &
  \multicolumn{1}{c}{-} &
  \multicolumn{1}{c|}{-} &
  \multicolumn{1}{c}{-} &
  - \\
SRF~\cite{10239230} & TITS 2023 &
  \multicolumn{1}{c}{82.4} &
  \multicolumn{1}{c}{97.5} &
  98.9 &
  \multicolumn{1}{c}{82.7} &
  \multicolumn{1}{c|}{95.1} &
  \multicolumn{1}{c}{77.9} &
  \multicolumn{1}{c|}{93.8} &
  \multicolumn{1}{c}{70.2} &
  90.6 \\ 
MART~\cite{9945658}$^{*}$ & TITS 2023 &
  \multicolumn{1}{c}{82.7} &
  \multicolumn{1}{c}{97.6} &
  98.7 &
  \multicolumn{1}{c}{83.8} &
  \multicolumn{1}{c|}{94.2} &
  \multicolumn{1}{c}{78.9} &
  \multicolumn{1}{c|}{93.9} &
  \multicolumn{1}{c}{70.8} &
  90.2 \\
BIDA~\cite{10013940} & TITS 2023 &
  \multicolumn{1}{c}{87.4} &
  \multicolumn{1}{c}{97.4} &
  98.5 &
  \multicolumn{1}{c}{-} &
  \multicolumn{1}{c|}{-} &
  \multicolumn{1}{c}{-} &
  \multicolumn{1}{c|}{-} &
  \multicolumn{1}{c}{-} &
  - \\ 
CLIP-ReID~\cite{li2023clip} & AAAI 2023 &
  \multicolumn{1}{c}{88.3} &
  \multicolumn{1}{c}{97.5} &
  98.0 &
  \multicolumn{1}{c}{-} &
  \multicolumn{1}{c|}{-} &
  \multicolumn{1}{c}{-} &
  \multicolumn{1}{c|}{-} &
  \multicolumn{1}{c}{-} &
  - \\ 
CLIP-ReID~\cite{li2023clip}$^{*}$ & AAAI 2023 &
  \multicolumn{1}{c}{91.7} &
  \multicolumn{1}{c}{98.0} &
  98.5 &
  \multicolumn{1}{c}{-} &
  \multicolumn{1}{c|}{-} &
  \multicolumn{1}{c}{-} &
  \multicolumn{1}{c|}{-} &
  \multicolumn{1}{c}{-} &
  - \\ 

MBR~\cite{10422175}& ITSC 2023 &
  \multicolumn{1}{c}{\underline{91.9}} &
  \multicolumn{1}{c}{\underline{98.2}} &
  98.4 &
  \multicolumn{1}{c}{{\underline{88.1}}} &
  \multicolumn{1}{c|}{96.3} &
  \multicolumn{1}{c}{\underline{83.8}} &
  \multicolumn{1}{c|}{95.1} &
  \multicolumn{1}{c}{\underline{77.4}} &
  92.4 \\ \hline
Baseline & &
  \multicolumn{1}{c}{86.7} &
  \multicolumn{1}{c}{96.8} &
   97.9&
  \multicolumn{1}{c}{84.9} &
  \multicolumn{1}{c|}{95.9} &
  \multicolumn{1}{c}{80.2} &
  \multicolumn{1}{c|}{94.8} &
  \multicolumn{1}{c}{72.9} &
   92.4\\ 
CLIP-SENet$^{*}$ & &
  \multicolumn{1}{c}{\textbf{92.9}} &
  \multicolumn{1}{c}{\textbf{98.7}} &
  \textbf{99.1} &
  \multicolumn{1}{c}{\textbf{89.1}} &
  \multicolumn{1}{c|}{\textbf{97.9}} &
  \multicolumn{1}{c}{\textbf{85.2}} &
  \multicolumn{1}{c|}{\textbf{97.0}} &
  \multicolumn{1}{c}{\textbf{79.5}} &
   \textbf{95.4}\\ \hline
\end{tabular}
}
\end{table*}

\section{Experiments}
\subsection{Dataset Analysis}
We evaluate our CLIP-SENet model on three challenging vehicle Re-ID benchmark datasets.

\textbf{VeRi-776 dataset}~\cite{liu2016deep} comprises more than 50,000 images 
of 776 unique vehicles, taken from 20 different camera perspectives with varying lighting conditions and orientations. It is divided into a training set with 576 identities and 37,778 images and a testing set with 200 identities and 11,579 images. Additionally, 1,678 images from the testing set serve as queries. The dataset provides labels for vehicle IDs, camera IDs, color IDs, and type IDs. Notably, Wang et al. ~\cite{Wang_2017_ICCV} annotated viewpoint information, including front, front side, rear side, and rear.

\textbf{VehicleID dataset}~\cite{liu2016deep2} consists of 221,763 images from 26,267 unique vehicles, captured by one camera from either front or back views. Half of the identities (13,164) are designated for training, while the other half for testing across different gallery sizes: Test800 (Small), Test1600 (Medium), and Test2400 (Large). Unlike other datasets, the VehicleID dataset contains only partial information on vehicle attributes.

\textbf{VeRi-Wild dataset}~\cite{lou2019veri} stands out as the largest vehicle Re-ID dataset, encompassing 174 camera views, 416,314 images, and 40,671 unique vehicle identities. Notably, it offers attribute labels for vehicle model, color, and type. The testing set is divided into three subsets, each with varying numbers of unique IDs: 3,000 (small), 5,000 (medium), and 10,000 (large). 

The details of these datasets are summarized in Table~\ref{table:dataset}. It is worth noting that the VehicleID dataset provides detailed annotations down to the brand of the vehicle type, whereas the other two datasets classify vehicles based on their appearance. 

\begin{table}[t]
\centering
\renewcommand{\arraystretch}{1.2}
\caption{Statistics of vehicle Re-ID datasets.}
\label{table:dataset}
\begin{tabular}{c|ccccc}
\hline
Dataset&Image &ID &Camera & Type& Color \\ \hline
VeRi-776 & 51,035 & 776 & 20 & 9 & 10 \\
VehicleID & 221,763 & 26,267 & 1 & 250 & 7 \\
VeRi-Wild & 416,314 & 40,671 & 174 & 8 & 11 \\
\hline
\end{tabular}
\end{table}

\begin{table}[t]
\centering
\renewcommand{\arraystretch}{1.2}
\caption{Comparison with state-of-the-art methods on VehicleID dataset.}
\label{table:vid}
\resizebox{\columnwidth}{!}{
\begin{tabular}{c|cc|cc|cc}
\hline
\multirow{2}{*}{Method} &
\multicolumn{2}{c|}{Small} &
\multicolumn{2}{c|}{Medium} &
\multicolumn{2}{c}{Large} \\ \cline{2-7} 
 & R1 & R5 & R1 & R5 & R1 & R5 \\ \hline  
PVEN~\cite{meng2020parsing} & 78.4 & 92.3 & 75.0 & 88.3 & 74.2 & 86.4 \\
SAN~\cite{qian2020stripe} & 79.7 & 94.3 & 78.4 & 91.3 & 75.6 & 88.3 \\
SAVER~\cite{khorramshahi2020devil}& 79.9 & 95.2 & 77.6 & 91.1 & 75.3 & 88.3 \\
CAL~\cite{Rao_2021_ICCV} & 82.5 & 94.7 & 78.2 & 91.0 & 75.1 & 88.5 \\
vehicleNet~\cite{zheng2020vehiclenet} & 83.6 & 96.8 & 81.3 & 93.6 & 79.4 & 92.0 \\
SOFCT~\cite{yu2023semantic}$^{*}$ & 84.5 & 96.8 & 80.9 & 95.2 & 78.7 & 93.7 \\
GiT~\cite{shen2023git}$^{*}$ & 84.6 & - & 80.5 & - & 77.9 & - \\
CLIP-ReID~\cite{li2023clip} & 85.2 & 97.1 & 80.7 & 94.3 & 78.7 & 92.3 \\
TransReID~\cite{He_2021_ICCV}$^{*}$ & 85.2 & 97.6 & - & - & - & - \\
CLIP-ReID$^{*}$ & 85.3 & 97.6 & 81.0 & 95.0 & 78.1 & 92.7 \\
ASSEN~\cite{li2022attribute} & 86.0 & 97.8 & \underline{84.5} & 96.0 & \underline{82.4} & 94.3 \\
MsKAT~\cite{9764648}$^{*}$& 86.3 & 97.4 & 81.8 & 95.5 & 79.5 & 93.9 \\
GSE-Net~\cite{10595990} & 86.4 & 97.7 & 84.5 & 96.4 & 81.8 & 93.5 \\
FastReID~\cite{he2023fastreid} & 86.6 & 97.9 & 82.9 & 96.0 & 80.6 & 93.9 \\
SRF~\cite{10239230}& 86.9 & 98.3 & 82.4 & \textbf{96.5} & 79.9 & 94.5 \\
SVRN~\cite{9837795}& 87.5 & - & 84.6 & - & 81.8 & - \\
BIDA~\cite{10013940} & 87.6 & \underline{98.5} & 83.9 & 96.3 & 80.6 & 93.8 \\
ANet~\cite{quispe2021attributenet}  & 87.9 & 97.8 & 82.8 & 96.2 & 80.5 & \underline{94.6} \\
HRCN~\cite{zhao2021heterogeneous} & 88.2 & 98.4 & 81.4 & \textbf{96.5} & 80.2 & 94.4 \\
MBR~\cite{10422175} & \underline{88.3} & - & - & - & - & - \\
CAN~\cite{sheng2023discriminative}$^{*}$ & \underline{88.3} & \underline{98.5} & 83.3 & \underline{96.4} & 80.5 & 94.5 \\
\hline
Baseline & 83.9 & 97.1 & 79.8 & 94.5 & 77.6 & 91.8 \\
CLIP-SENet$^{*}$ & \textbf{90.4} & \textbf{98.7} & \textbf{85.5} & \textbf{96.5} & \textbf{82.7} & \textbf{94.8} \\
\hline
\end{tabular}
}
\end{table}

\begin{table}[t]
\centering
\renewcommand{\arraystretch}{1.2}
\caption{Ablation study of the components in CLIP-SENet. The CV represents the incorporation of additional camera and viewpoint information.}
\label{table:1}
\begin{tabular}{c|c|c|c|c|c}\hline
SEM & AFEM & CV & mAP (\%) & Rank-1 (\%) & Rank-5 (\%) \\
\hline
- & - & - & 86.7 & 96.8 & 97.9 \\
\checkmark & - &  - & 87.3 & 97.0 & 97.6\\
\checkmark & \checkmark & - & 91.4 & 97.6 & 98.5 \\
- & - &  \checkmark & 86.8 & 97.1 & 98.1\\
\checkmark & - & \checkmark & 86.4 & 96.1 & 97.3 \\
\checkmark & \checkmark& \checkmark &\textbf{92.9} & \textbf{98.7} & \textbf{99.1}\\
\hline
\end{tabular}
\end{table}

\begin{table}[t]
  \centering
  \renewcommand{\arraystretch}{1.2}
    \caption{Ablation study on the number of groups in AFEM.}
    \label{table:2}
   \begin{tabular}{c|c|c|c}
    \hline
    groups & mAP (\%) & Rank-1 (\%) & Rank-5 (\%) \\
    \hline
    4 & 90.2 & 97.2 & 98.1 \\
    8 & 91.1 & 97.6 & 98.1 \\
    16 & 91.7 & 98.2 & 98.9 \\
    32 &\textbf{92.9} & \textbf{98.7} & \textbf{99.1}\\
    64 & 90.3 & 97.9 & 98.7 \\
    128 & 90.6 & 97.9 & 98.7 \\
    \hline
    \end{tabular}
\end{table}

\begin{table}[t]
  \label{table:4}
  \centering
  \renewcommand{\arraystretch}{1.2}
      \caption{Ablation Study on CNN Backbone Network.}
    \label{table:3}
   \begin{tabular}{c|c|c|c}
    \hline
    Backbone & mAP (\%) & Rank-1 (\%) & Rank-5 (\%) \\
    \hline
    ResNet50 & 91.1 & 97.4 & 98.6 \\
    ResNet101 & 91.9 & 98.2 & 98.9 \\
    SE-ResNet101 & 91.0 & 97.8 & 98.7 \\
    ResNext101 & \textbf{92.9} & \textbf{98.7} & \textbf{99.1}\\
    \hline
    \end{tabular}
\end{table}

\begin{figure}[!t]
\centering
\includegraphics[width=0.85\linewidth]{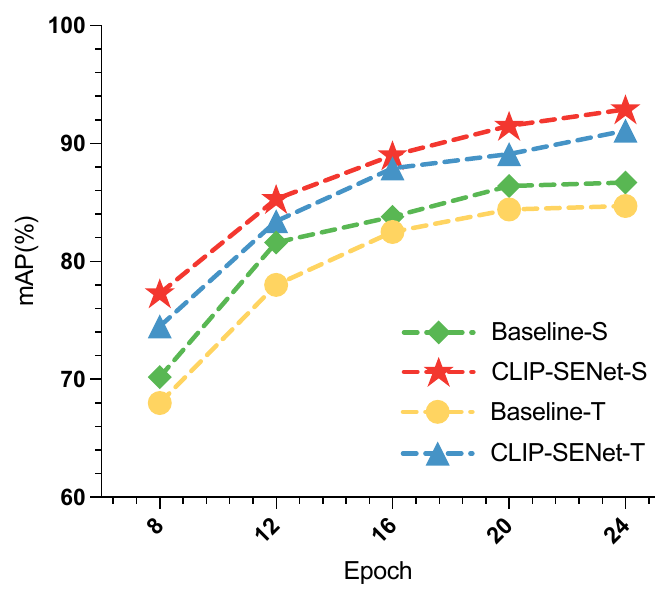}
\caption{Ablation study on loss function. ``-S" means using SupCon loss to guide the training and ``-T" means using Triplet loss to guide the training.}
\label{fig:loss}
\end{figure}

\subsection{Implement Details}
\subsubsection{Training details}In our experimental setup, we employ a ResNext101 with IBN structure as our CNN Backbone, combined with image encoder from TinyCLIP based on ViT-B/32. Our image preprocessing involves resizing the image size to 320x320 and accompanying by various data augmentation techniques. For model optimization, we adopted the ADAM optimizer~\cite{schroff2015facenet} and integrated a cosine annealing scheduler, starting with an initial learning rate of 5e-4. The batch size was set to 128, with each batch containing samples from 16 different vehicle identities, each represented by 8 images. The number of epochs was configured to 24. 
For extensive datasets such as VehicleID and VeRi-Wild, adjustments were made to accommodate the larger data volume, setting the training duration to 120 epochs and using WarmupMultiStepLR to optimize the learning rate, initialized to 3.5e-4 for effective data fitting.
Our entire experimental process was conducted on an NVIDIA A40 GPU, with the global seed set to 3407 to ensure the reproducibility of our experiments.

\subsubsection{Evaluation}In line with common practices, we evaluated the performance of model using mean Average Precision (mAP) and Cumulative Matching Characteristics at Rank 1 (Rank-1) and Rank 5 (Rank-5) as primary metrics. 
Throughout this study, we employed re-ranking technology as a post-processing step only for the VeRi-776 dataset.

\subsubsection{Compared Methods}We compare our method with some state-of-the-art approaches, which can be categorized into three main groups.

\textit{a) Global feature embedding based method:} Models that operate based on global features. \textit{E.g.}, GCN-based global structure embedded network (GSE-Net)~\cite{10595990}, FastReID~\cite{he2023fastreid}, TransReID~\cite{He_2021_ICCV}, Spatially-Regularized Features (SRF)~\cite{10239230}, Heterogeneous Relational Complement Network (HRCN)~\cite{zhao2021heterogeneous}, vehicleNet~\cite{zheng2020vehiclenet}, Salience-Navigated Vehicle Re-identification Network (SVRN)~\cite{9837795}, StrongBaseline~\cite{huynh2021strong}, Multi-Branch Representation(MBR)~\cite{10422175}. 

\textit{b) Fine-grained based method:} Models that work on local and fine-grained detials. \textit{E.g.}, Counterfactual Attention Learning(CAL)~\cite{Rao_2021_ICCV}, parsing-based view-aware embedding network
(PVEN)~\cite{meng2020parsing}, Dual cross- attention learning (DCAL)~\cite{zhu2022dual}, graph interactive transformer (GiT)~\cite{shen2023git}, Self-supervised Attention for Vehicle Re-identification (SAVER)~\cite{khorramshahi2020devil}, Mask-Aware Reasoning Transformer (MART)~\cite{9945658}, Co-occurrence Attention Net (CAN)~\cite{sheng2023discriminative}.

\textit{c) Attribute based method:} Models enhanced by semantic or attribute-based information. \textit{E.g.}, stripebased and attribute-aware deep convolutional neural network (SAN)~\cite{qian2020stripe}, Semantic-oriented feature coupling transformer (SOFCT)~\cite{yu2023semantic}, Attributenet (ANet)~\cite{quispe2021attributenet}, Attribute and state guided structural embedding network (ASSEN)~\cite{li2022attribute}, Multi-scale Knowledge-Aware Transformer (MsKAT)~\cite{9764648}, Bi-level implicit semantic data augmentation (BIDA)~\cite{10013940}, CLIP-ReID~\cite{li2023clip}.

\begin{figure*}[t]
\centering
\subfloat[baseline]{\includegraphics[width=0.33\linewidth]{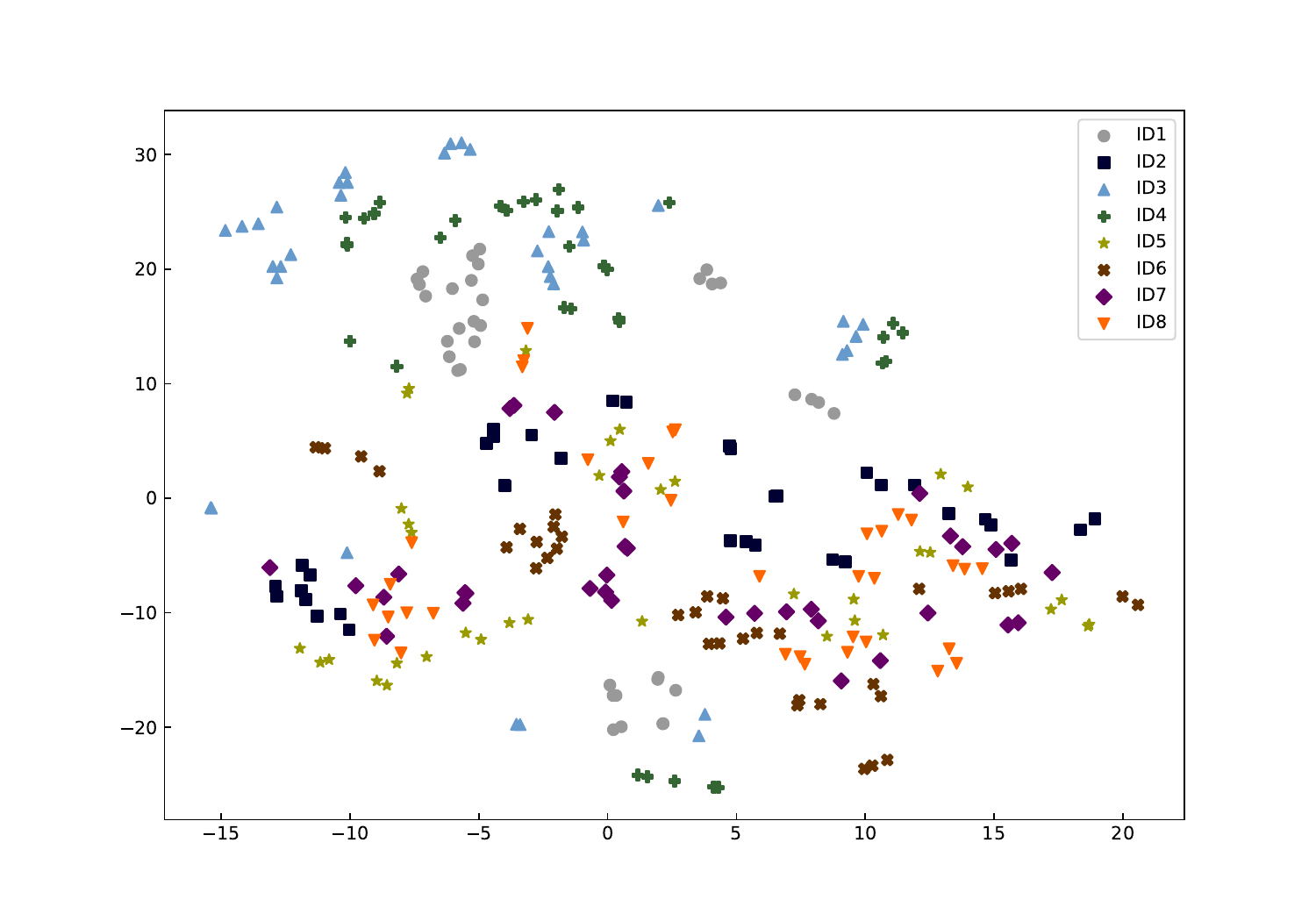}%
\label{fig_a}}
\hfil
\subfloat[baseline+SEM]{\includegraphics[width=0.33\linewidth]{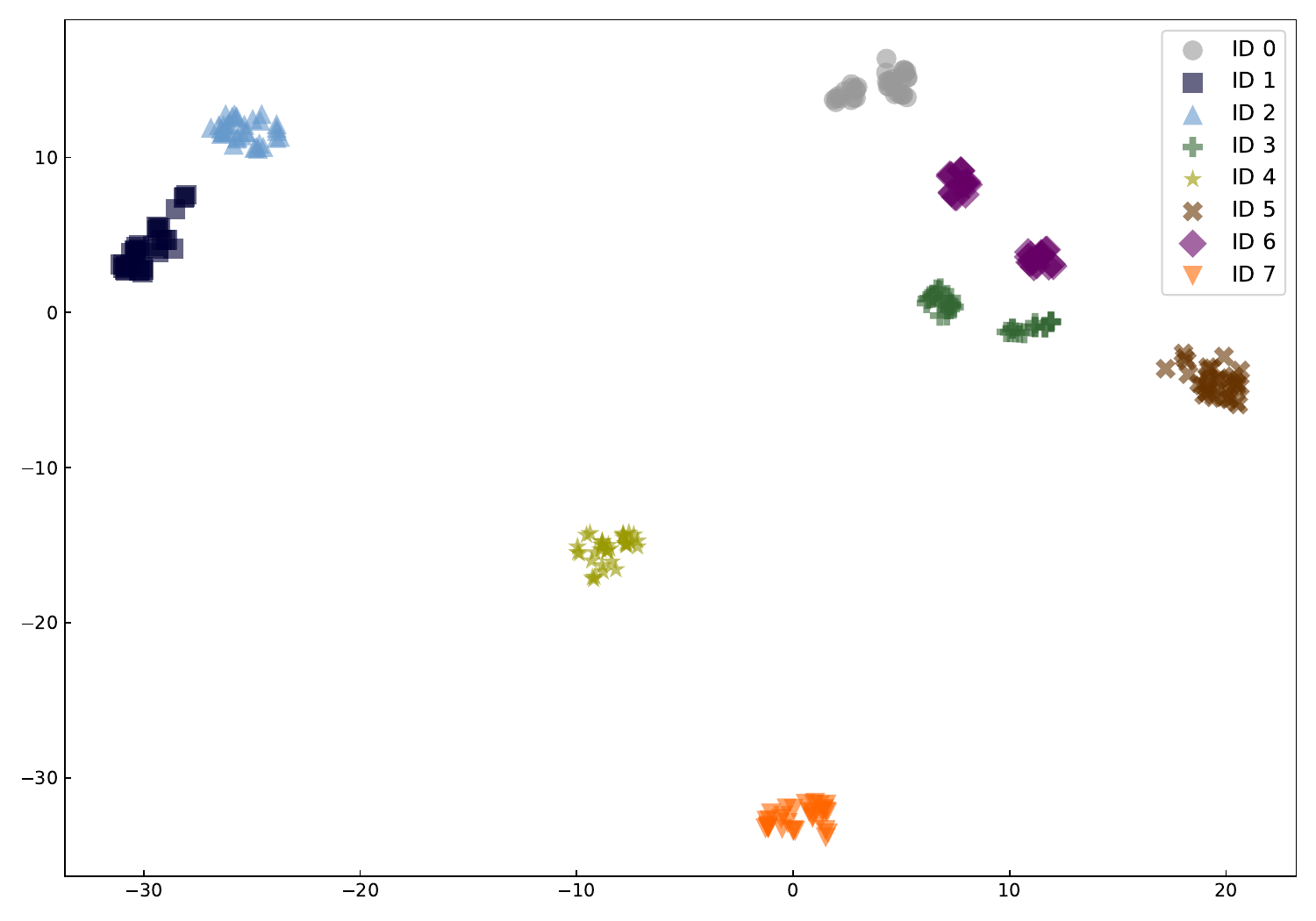}%
\label{fig_b}}
\hfil
\subfloat[baseline+SEM+AFEM]{\includegraphics[width=0.33\linewidth]{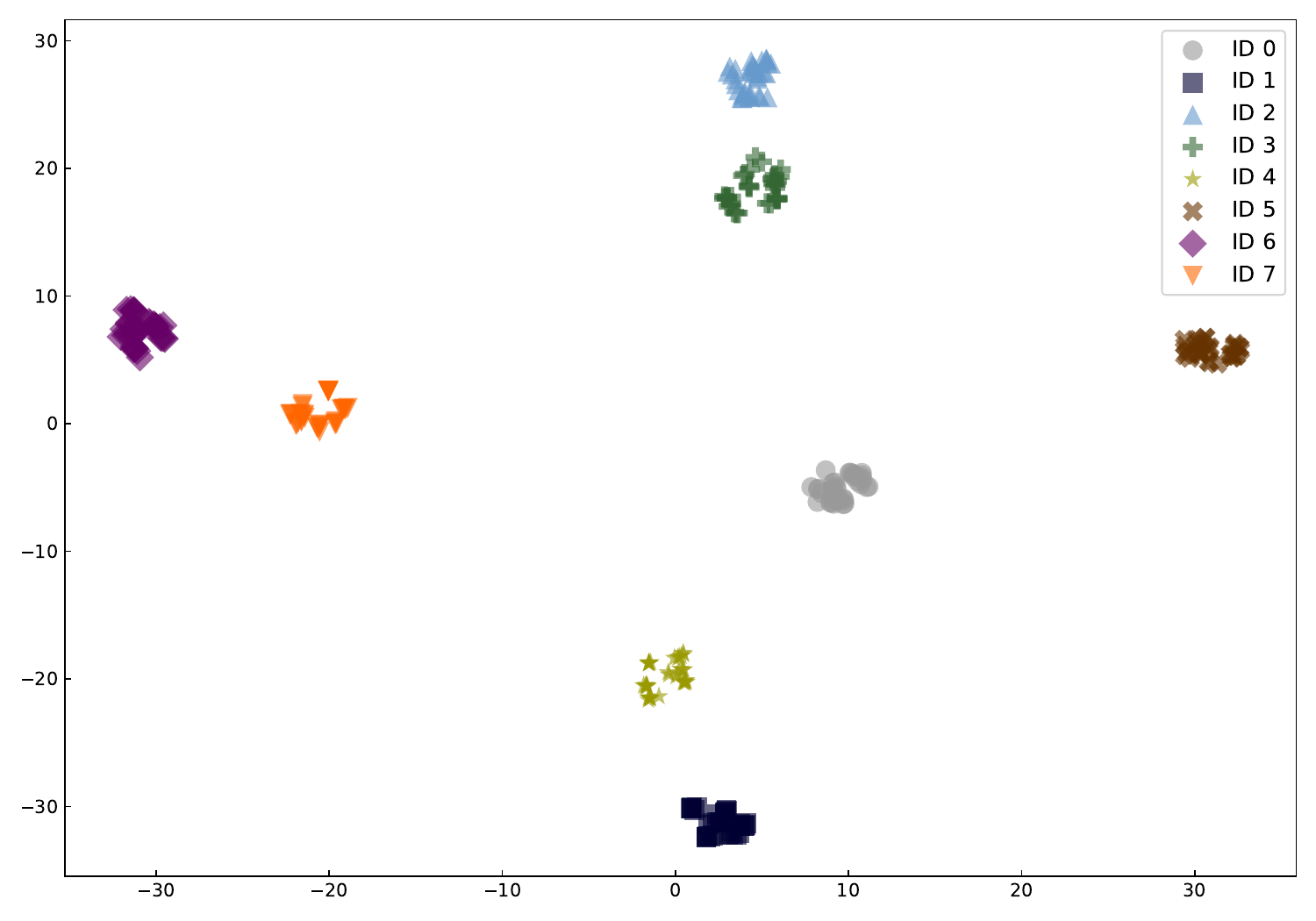}%
\label{fig_c}}

\subfloat[baseline+CV]{\includegraphics[width=0.33\linewidth]{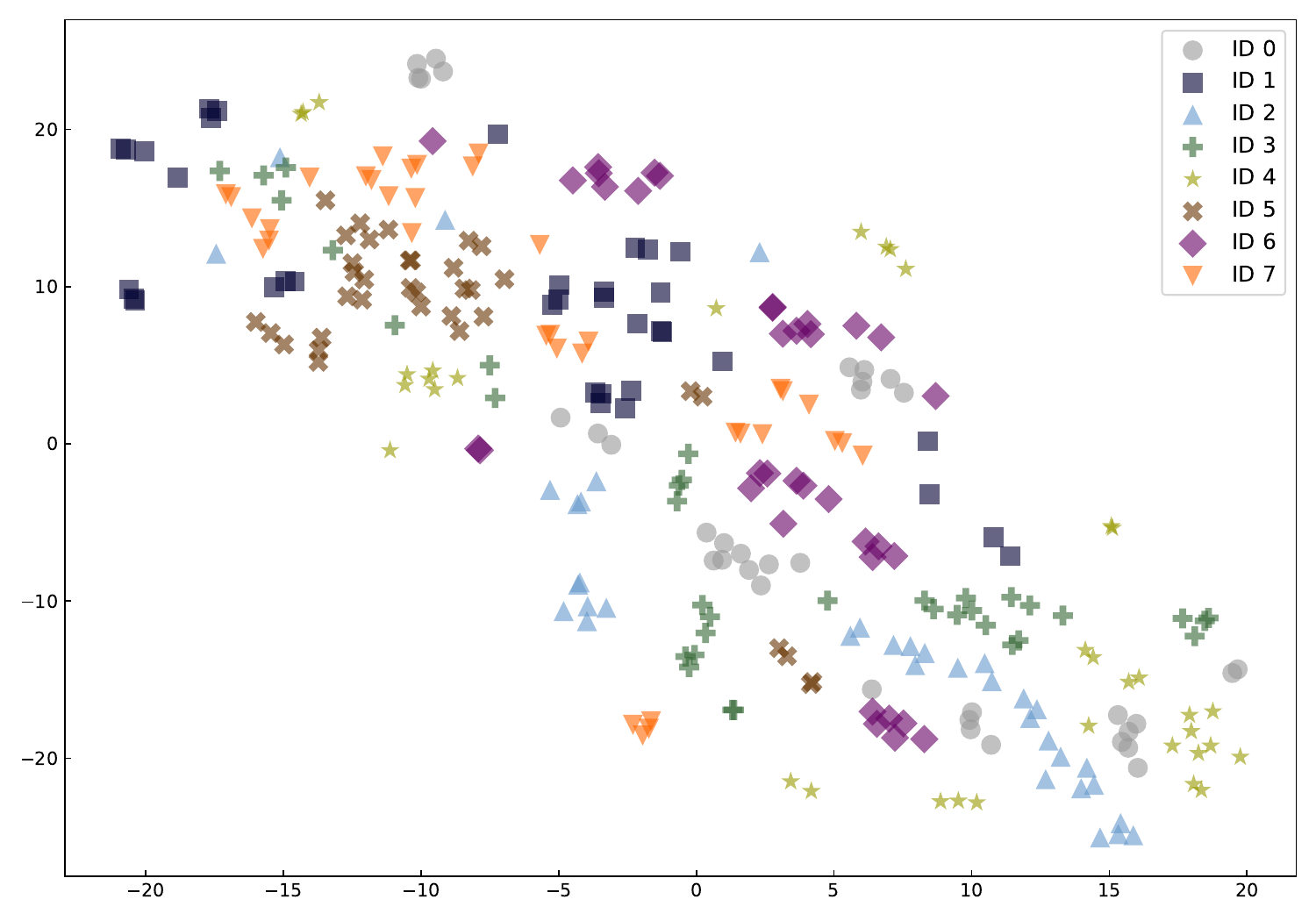}%
\label{fig_d}}
\hfil
\subfloat[baseline+SEM+CV]{\includegraphics[width=0.33\linewidth]{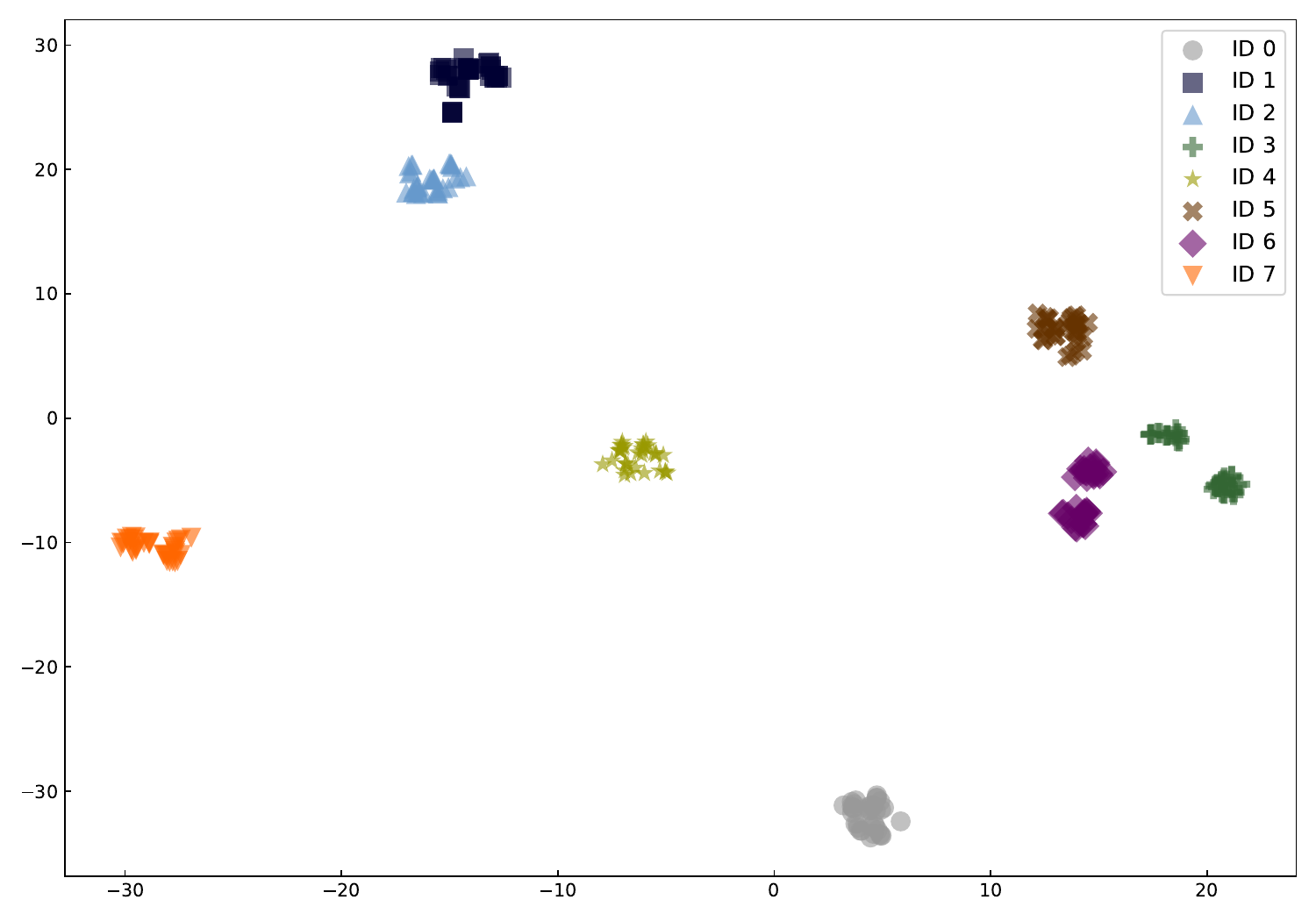}%
\label{fig_e}}
\hfil
\subfloat[baseline+SEM+AFEM+CV]{\includegraphics[width=0.33\linewidth]{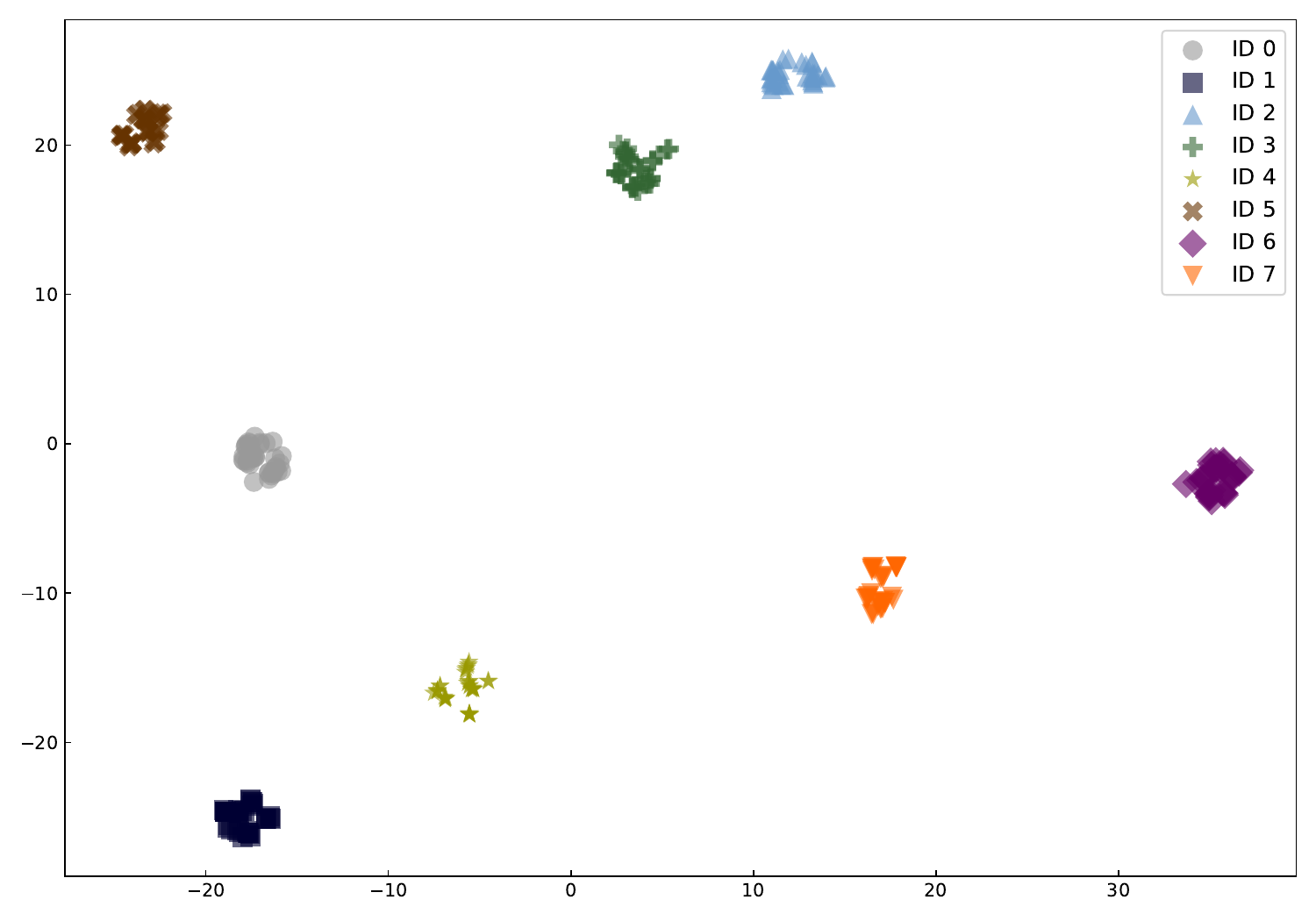}%
\label{fig_f}}
\caption{T-SNE~\cite{van2008visualizing} visualization of features extracted by the model. Randomly selected 36 images from each of 8 vehicle IDs on the VeRi-776 dataset are represented with different colors.}
\label{fig:t-sne}
\end{figure*}

\subsection{Comparison with State-of-the-Art Methods.}
We conducted a comprehensive comparison of CLIP-SENet with state-of-the-art methods on three benchmark datasets in vehicle Re-ID, detailed in Table~\ref{table:5} and Table~\ref{table:vid}.

\subsubsection{Evaluation Result on VeRi-776} Our model achieved 92.9\% mAP and 98.7\% Rank-1, representing a significant improvement of 1.2\% in mAP and 0.7\% in Rank-1 over the CLIP-based method CLIP-ReID (ViT-based). This demonstrates the effectiveness of our enhancements to the image encoder from TinyCLIP. By discarding semantic alignment, our method of enhancing the original visual semantic features with fine-grained details proves effective. Compared with the fine-grained method CAN, our approach significantly improves mAP by 9.5\% and Rank-1 by 1.1\%. This highlights the advantages of our model, which fully considers both semantic and fine-grained features. Furthermore, our CLIP-SENet notably surpassed the previously best-performing method on the VeRi-776 dataset, MBR, by 1\% in mAP and 0.5\% in Rank-1.

\subsubsection{Evaluation Result on VehicleID}
Our model also achieved competitive results on the VehicleID dataset, with 90.4\% Rank-1 and 98.7\% Rank-5 on the small test set, significantly surpassing previous methods based on global feature embedding. For vehicle datasets captured from only two viewpoints by a single camera, the advantages of fine-grained and attribute-based methods become evident. Compared to the fine-grained approach CAN, our CLIP-SENet improves Rank-1 accuracy by 2.1\%, 2.2\%, and 2.2\% on the small, medium, and large test sets, respectively.
Additionally, our model maintained a notable advantage over the ViT-based CLIP-ReID. Specifically, on the VehicleID test set, our model improved Rank-1 accuracy by 5.1\% (small), 4.5\% (medium), and 4.6\% (large) compared to CLIP-ReID.

\subsubsection{Evaluation Result on VeRi-Wild}
On the larger and more challenging VeRi-Wild dataset, our CLIP-SENet consistently demonstrates performance advantages. Compared to the best attribute-enhanced method, ANet, our model improves Rank-1 performance by 3.2\%, 2.7\%, and 3.6\% on the small, medium, and large test sets, respectively. When compared to the best fine-grained method, CAN, our model shows improvements of 1.6\%, 2\%, and 2.2\%. Additionally, compared to the best global feature-based method, MBR, our model improves by 1\%, 1.4\%, and 2.1\%. This further proves the advantage of CLIP-SENet by combining semantic and fine-grained features.

\subsection{Ablation Study}
To validate the effectiveness of our method, under fixed training conditions, we used the VeRi-776 dataset as our benchmark dataset and set a global random seed. 

\subsubsection{Baseline}
We only employ ResNeXt101-IBN as the CNN Backbone without CLIP and optimize it under the guidance of CE loss and SupCon Loss, establishing this configuration as our baseline. Our baseline model achieves a performance of 86.7\% mAP, 96.8\% Rank-1 and 97.9\%Rank-5 on the VeRi-776 dataset.

\subsubsection{Ablation Study on Components in CLIP-SENet}
We progressively added the SEM and AFEM to construct our CLIP-SENet and verified the effectiveness of each component, as shown in Table~\ref{table:1}.
Upon adding the SEM alone, our model achieves 87.3\% mAP and 97.0\% Rank-1, representing only a 0.6\% improvement in mAP compared to the baseline. This suggests that the simple concatenation and linear projection of raw semantic features do not yield a significant improvement in performance. This is likely because the raw semantic information contains irrelevant noise that impairs vehicle Re-ID discrimination.
To address this, we incorporated AFEM to refine the semantic features. Under the same conditions, adding AFEM resulted in an mAP of 91.4\%, while Rank-1 remained at 97.6\%. Like many methods applied to VeRi-776, when our model incorporates camera and viewpoint information, its performance further improves to 92.9\% mAP and 98.7\% Rank-1.
To verify the impact of CV information, we conducted comparative experiments each time we added a component. It is evident that simply adding CV information resulted in limited performance improvement. This further verifies the effectiveness of AFEM. 

We use t-SNE visualization to intuitively demonstrate the impact of incrementally adding each component on the data distribution, as shown in Fig.~\ref{fig:t-sne}. From a horizontal perspective, we can observe that as we gradually add components, the features belonging to each ID become increasingly compact and clustered. From a vertical perspective, we can conclude that the inclusion of CV has a significantly positive effect on datasets, effectively reducing the intra-class distance.

\begin{figure}[t]
\centering
\includegraphics[width=\columnwidth]{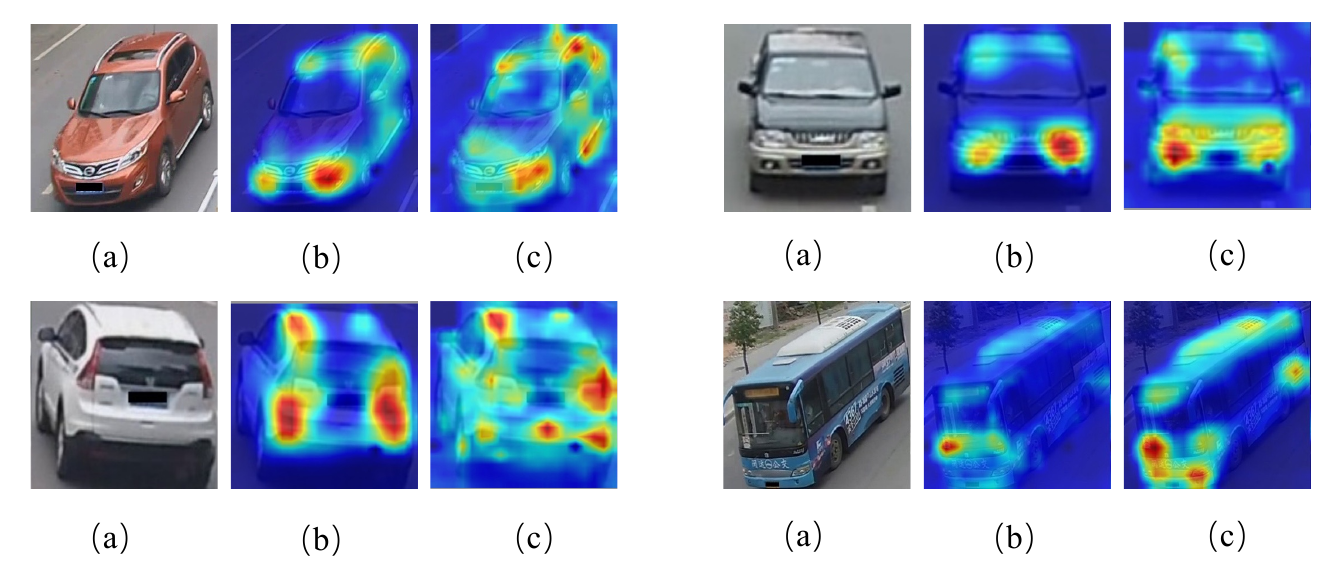}
\caption{Activation map visualization. (a) Input images, (b) our Baseline, (c) our CLIP-SENet.}
\label{fig:actmap}
\end{figure}

\subsubsection{Ablation Study on Loss Function}
To validate the effectiveness of the SupCon Loss, we conducted experiments on both the baseline and CLIP-SENet, comparing the results with the commonly used triplet loss. As shown in Fig.~\ref{fig:loss}, both the baseline and CLIP-SENet exhibit performance improvement when using the SupCon Loss. 

\subsubsection{Ablation Study on Hyperparameters}
The number of groups in AFEM is a critical hyperparameter, representing how many semantic attributes need attention. 
Through our experiments with varying group numbers, as shown in Table~\ref{table:2}, we observed that the model's performance initially improves as the number of groups increases but eventually declines. 
This finding highlights the importance of balancing the granularity of semantic attribute grouping. Having too few groups may lead to insufficient capture of semantic details, while too many groups might introduce noise and redundancy, negatively impacting the performance of model. Therefore, setting the number of groups to 32 strikes a balance, enhancing the discriminative power and effectiveness of our model.

\subsubsection{Ablation Study on CNN Backbone Network}
Leveraging the pre-trained weights provided by IBN-Net, we selected ResNet50-IBN, ResNet101-IBN, SE-ResNet101-IBN, and ResNeXt101-IBN for ablation experiments. Table~\ref{table:3} shows that ResNet101-IBN outperforms ResNet50-IBN, with a 0.8\% increase in mAP due to its larger size. However, SE-ResNet101-IBN performs worse than ResNet101-IBN and even lower in mAP than ResNet50-IBN, likely because the information extracted by the SE structure does not integrate well with the information extracted by CLIP. The best-performing backbone is ResNeXt101-IBN, with a 1\% increase in mAP and a 0.5\% increase in Rank-1 over ResNet101-IBN.

\begin{figure}[!t]
\centering
\includegraphics[width=\columnwidth]{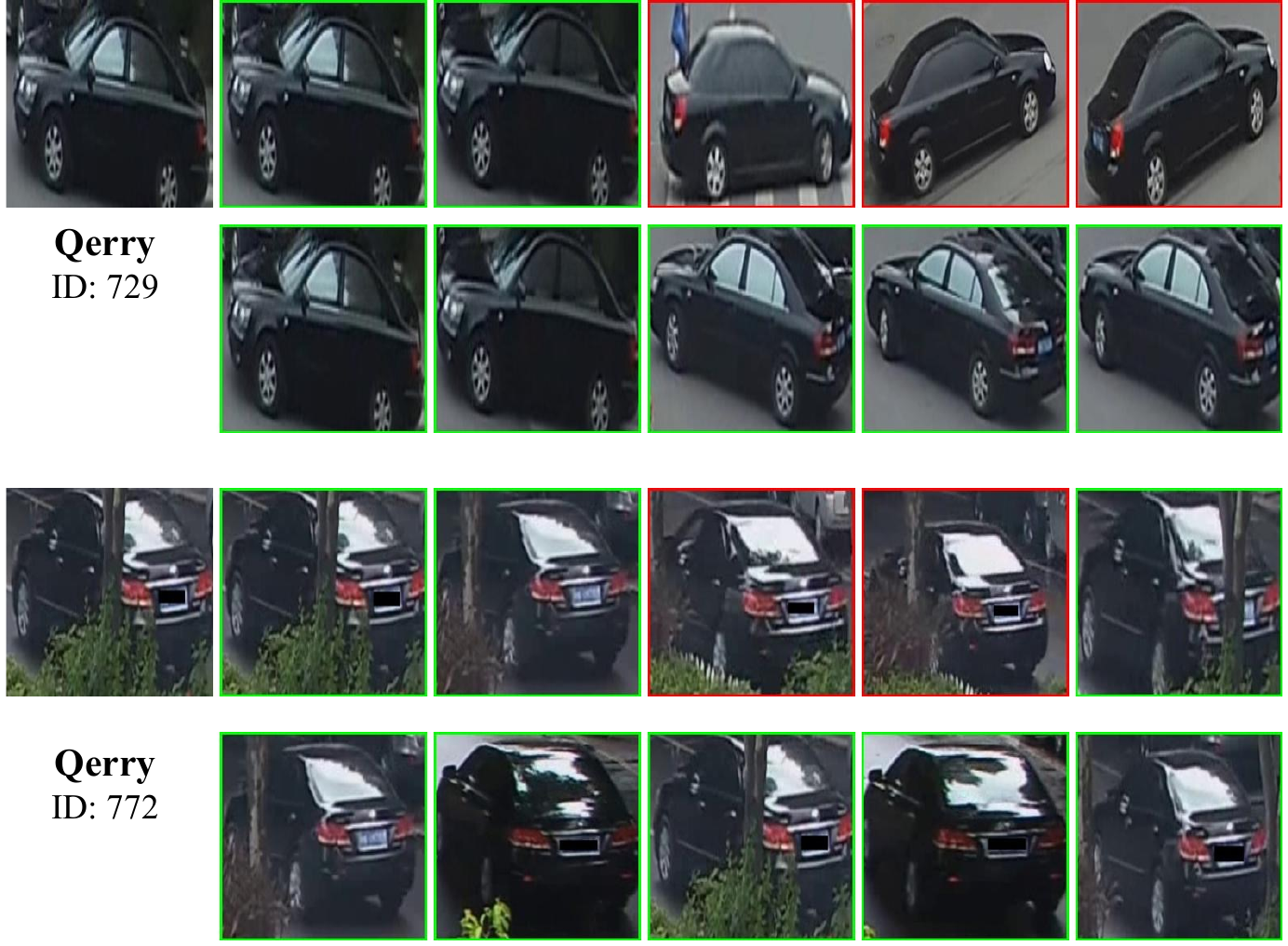}
\caption{Rank-5 visualization examples on the VeRi-776 dataset. The first row illustrates the query image and top five gallery images retrieved by the Baseline model, while the second row presents those obtained by CLIP-SENet. The green boxes represent true positive samples, accurately matching the query vehicle, whereas red boxes denote false positive samples, incorrectly identified as matches.}
\label{fig:ranklist}
\end{figure}

\subsection{Visualization of CLIP-SENet}
To demonstrate the powerful feature representation capabilities of CLIP-SENet on the VeRi-776 dataset and its improvements over baseline models, we employed various visualization techniques for a more intuitive description. 

The activation map visualization experiments highlight the image regions that the model focuses on the most. As shown in Fig. \ref{fig:actmap}, for each provided example, CLIP-SENet clearly concentrates on more comprehensive areas compared to the baseline, indicating a higher attention to relevant features. 

Moreover, the efficacy of CLIP-SENet in practical vehicle Re-ID scenarios was evaluated through Rank-5 list visualizations. This involved comparing the query images and their corresponding ranking lists generated by both the baseline and CLIP-SENet models, as depicted in Fig.~\ref{fig:ranklist}. Statistical analysis of retrieval accuracy revealed that, while the baseline model maintained a degree of robustness, it exhibited limitations in generating precise ranking lists when processing vehicles with closely resembling features. Conversely, CLIP-SENet consistently outperformed the baseline, demonstrating exceptional precision in generating ranking lists. This not only highlights the robust discriminative ability of CLIP-SENet but also its enhanced capability to accurately navigate the intricate challenges posed by similar vehicle appearances.

\section{Conclusion}
In this work, we investigate the potential of using the image encoder from TinyCLIP exclusively for extracting vehicle semantic attribute information. To address the shortcomings in attribute feature extraction within existing vehicle Re-ID frameworks, we introduce CLIP-SENet. First, we utilize image encoder to extract raw semantic attributes, leveraging its powerful zero-shot learning capabilities to eliminate the need for manual annotation. Next, we apply the adaptive fine-grained enhancement module to refine these raw semantic features, enabling adaptive weighting of semantic attributes for each vehicle. Finally, we fuse the refined semantic attributes with vehicle appearance features extracted by a CNN backbone, creating a robust Re-ID feature representation. Compared to previous attribute-based and CLIP-based methods, CLIP-SENet overcomes the reliance on additional text annotations and effectively enhances semantic representation. We hope that our research will push the field of vehicle Re-ID in a more promising direction.


\bibliographystyle{IEEEtran}
\bibliography{refer}
%


\vspace{-33pt}
\begin{IEEEbiography}[{\includegraphics[width=1in,height=1.25in,clip,keepaspectratio]{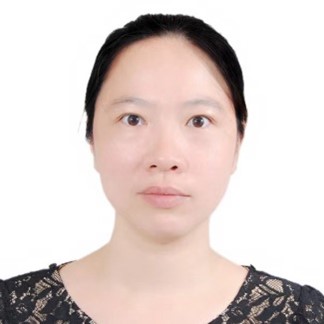}}]{Liping Lu} received the B.E. degree in industrial automation control and the M.S. degree in automation control theory from the Wuhan University of Technology, Wuhan, China, in 1996 and 1999, respectively, and the Ph.D. degree in computer science from the Institute National Polytechique de Lorraine, Nancy, France, in 2006. She is currently an Associate Professor with the School of Computer Science and Artificial Intelligence, Wuhan University of Technology, focusing on the research of autonomous vehicles and vehicular ad hoc networks.
\end{IEEEbiography}

\begin{IEEEbiography}[{\includegraphics[width=1in,height=1.25in,clip,keepaspectratio]{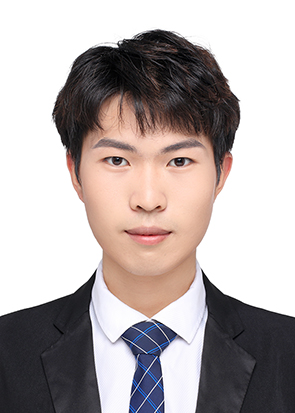}}]{Zihao Fu} received the B.S. degree in software
engineering from the Qingdao University, Qingdao, China, in 2022. He is currently pursuing the
M.S. degree with Wuhan University of Technology,
Wuhan, China. His research interests are focusing on
the research of computer vision based deep learning and vehicle
re-identification. 
\end{IEEEbiography}

\begin{IEEEbiography}[{\includegraphics[width=1in,height=1.25in,clip,keepaspectratio]{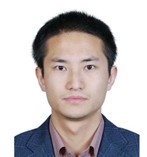}}]{Duanfeng Chu} (Member, IEEE) received the B.E.
and Ph.D. degrees in mechanical engineering from
the Wuhan University of Technology, Wuhan, China,
in 2005 and 2010, respectively.
He has visited California PATH, University of
California at Berkeley, Berkeley, USA, and the
Department of Mechanical and Aerospace Engineering, The Ohio State University, USA, in 2009 and
2017, respectively. He is currently a Professor
with the Intelligent Transportation Systems Research
Center, Wuhan University of Technology, focusing
on the research of automated and connected vehicle and intelligent transportation systems.
Dr. Chu has been a reviewer for several international journals and conferences in the field of connected and autonomous vehicles.
\end{IEEEbiography}

\begin{IEEEbiography}[{\includegraphics[width=1in,height=1.25in,clip,keepaspectratio]{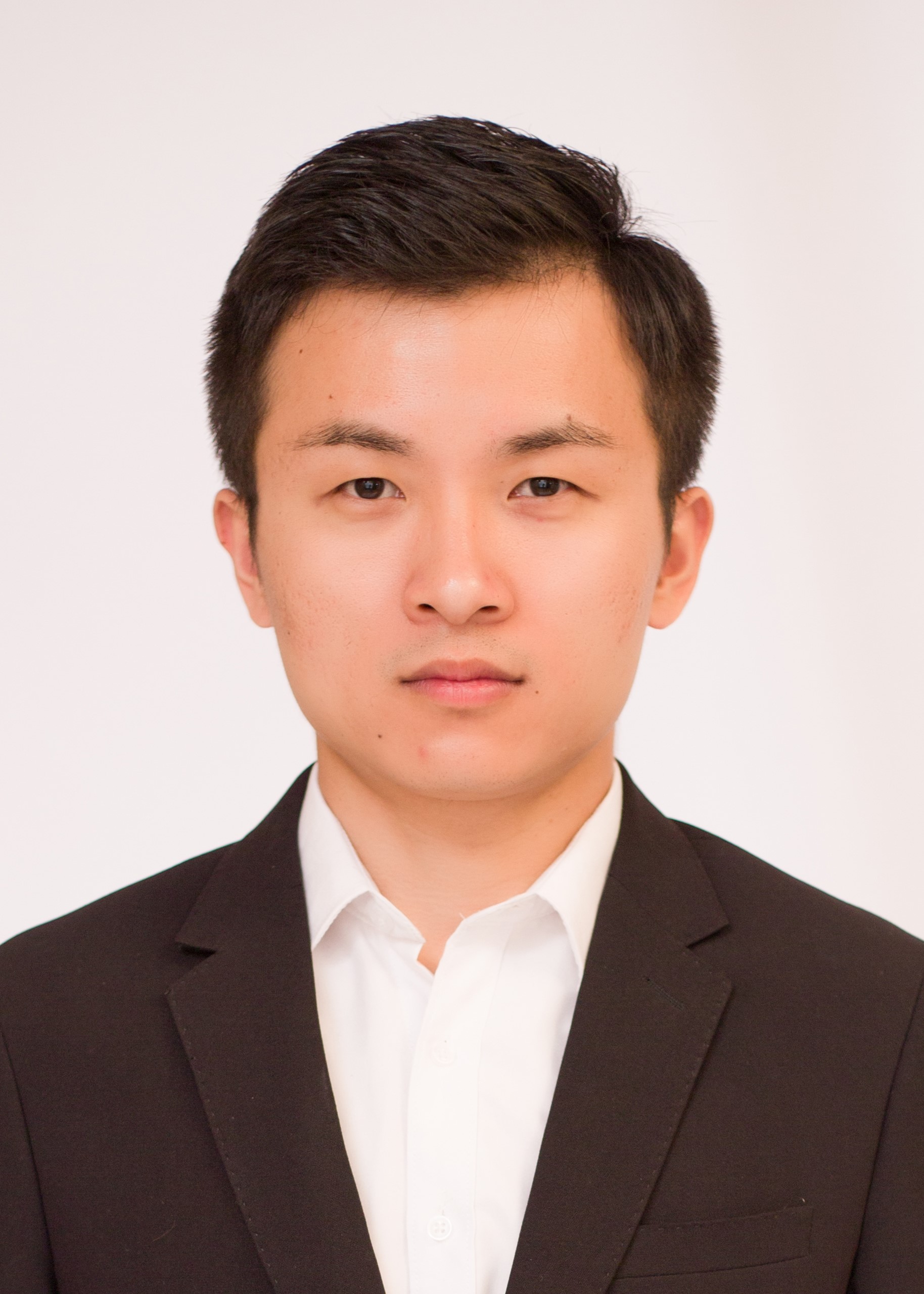}}]{Wei Wang}
is currently a post-doctoral candidate at the School of Cyber Science and Technology, Shenzhen Campus of Sun Yat-sen university, Shenzhen, China. He received the Ph.D. degree at the School of Software Technology, Dalian University of Technology, Dalian, China, in 2022. He received the M.S. degree at the School of Computer Science and Technology from the Anhui University, Hefei, China, in 2018. He received the B.S. degree at the School of Science from the Anhui Agricultural University, Hefei, China, in 2015. His major research interests include  transfer learning, zero-shot learning, deep learning, etc.
\end{IEEEbiography}

\begin{IEEEbiography}[{\includegraphics[width=1in,height=1.25in,clip,keepaspectratio]{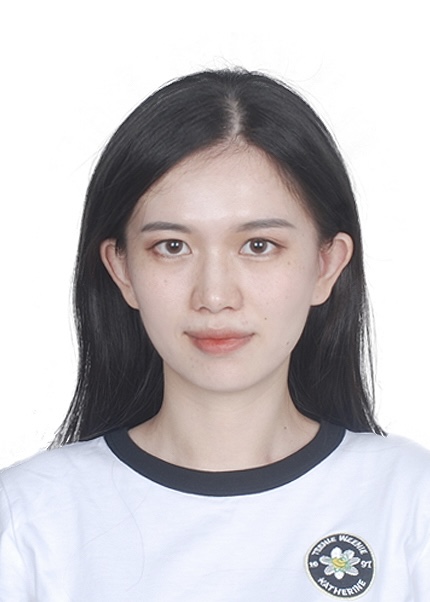}}]{Bingrong Xu} (Member, IEEE) received the B.Eng. degree from the School of Automation, Wuhan University of Technology, Wuhan, China, in 2015, and received the Ph.D. degree with the School of Artificial Intelligence and Automation, Huazhong University of Science and Technology, Wuhan, China, in 2021. She is currently an associate professor with the School of Automation, Wuhan University of Technology. Her current research interests include image processing, transfer learning, multi-view learning, and low-rank representation.
\end{IEEEbiography}


\vfill

\end{document}